\documentclass{myifcolog}

\usepackage[english]{babel}
\usepackage[utf8x]{inputenc}
\usepackage[T1]{fontenc}
\usepackage{cite}
\usepackage{xcolor}
\usepackage{bbold}
\usepackage{subfig}
\usepackage{enumitem}
\usepackage{comment}
\usepackage{float}

\usepackage{amsmath}
\usepackage{amssymb}
\usepackage{graphicx}
\usepackage[colorinlistoftodos]{todonotes}
\usepackage{physics}
\usepackage{tensor}
\usepackage{mathtools}
\usepackage{proof} 
\usepackage{prftree}
\usepackage{booktabs}
\usepackage{stmaryrd}
\usepackage{todonotes}
\usepackage{comment}

\newcommand{\cut}[1]{} 
\newcommand{\bs}{\backslash}
\newcommand{\nd}[2]{#1\vdash #2}
\newcommand{\Dg}[1]{\llbracket #1 \rrbracket_g}
\newcommand{\Spc}[1]{\lceil #1 \rceil}

\title{Density Matrices with metric for Derivational Ambiguity}

\titlerunning{Density Matrices with Metric for Derivational Ambiguity}

\titlethanks{}

\addauthor{Adriana D. Correia
\footnote{Corresponding author: a.duartecorreia@uu.nl}}{Institute for Theoretical Physics and Center for Complex Systems Studies, Utrecht University, P.O. Box 80.089, 3508 TB, Utrecht, The Netherlands.} 

\addauthor{Michael Moortgat}{Utrecht Institute of Linguistics OTS and Center for Complex Systems Studies, Utrecht University, 3512 JK, Utrecht, The Netherlands}

\addauthor{Henk T. C. Stoof}{Institute for Theoretical Physics and Center for Complex Systems Studies, Utrecht University, P.O. Box 80.089, 3508 TB, Utrecht, The Netherlands.}

\authorrunning{Correia, Moortgat and Stoof}

\begin{document}
\maketitle

\begin{abstract}

Recent work on vector-based compositional natural language semantics has proposed the use of density matrices to model lexical ambiguity and (graded) entailment (e.g. Piedeleu et al 2015, Bankova et al 2019, Sadrzadeh et al 2018). Ambiguous word meanings, in this work, are represented as mixed states, and the compositional interpretation of phrases out of their constituent parts takes the form of a strongly monoidal functor sending the derivational morphisms of a pregroup syntax to linear maps in FdHilb.

Our aims in this paper are threefold. Firstly, we replace the pregroup front end by a Lambek categorial grammar with directional implications expressing a word's selectional requirements. By the Curry-Howard correspondence, the derivations of the grammar's type logic are associated with terms of the (ordered) linear lambda calculus; these terms can be read as programs for compositional meaning assembly with density matrices as the target semantic spaces. Secondly, we extend on the existing literature and introduce a symmetric, nondegenerate bilinear form called a "metric" that defines a canonical isomorphism between a vector space and its dual, allowing us to keep a distinction between left and right implication. Thirdly, we use this metric to define density matrix spaces in a directional form, modeling the ubiquitous derivational ambiguity of natural language syntax, and show how this allows an integrated treatment of lexical and derivational forms of ambiguity controlled at the level of the interpretation.

\end{abstract}

\section{Introduction}
Semantic representations of language using vector spaces are an increasingly popular approach to automate natural language processing, with early comprehensive accounts given in \cite{mitchell2010composition, clark2015vector}. This idea has found several implementations, both theoretically and computationally. On the theoretical side, the principle of compositionality \cite{LoLacompositionality}
states that the meaning of a complex expression can be computed from the meaning of its simpler building blocks and the rules used to assemble them. On the computational side, the distributional hypothesis \cite{harris1954distributional} asserts that a meaning of a word is adequately represented by looking at what words most often appear next to it. Joining these two approaches, a distributional compositional categorical (DisCoCat) model of meaning has been proposed \cite{coecke2010mathematical}, mapping the pregroup algebra of syntax to vectors spaces with tensor operations, by functorialy relating the properties of the categories that describe those structures, allowing one to interpret compositionality in a grammar-driven manner using data-extracted representations of words that are in principle agnostic to grammar. This method has been shown to give good results when used to compare meanings of complex expressions and with human judgements \cite{grefenstette2011experimental}. Developments in the computation of these vectors that use machine learning algorithms \cite{mikolov2013distributed} provide representations of words that start deviating from the count-based models. However, each model still provides a singular vector embedding for each word, which allows the DisCoCat model to be applied with some positive results \cite{wijnholds2019evaluating}.

The principal limitation of these embeddings, designated \textit{static} embeddings, is that it provides the same word representation independently of context. This hide polysemy, or even subtler gradations in meaning. Using the DisCoCat framework, this issue has been tackled using density matrices to describe lexical ambiguity \cite{piedeleu2014ambiguity,DBLP:journals/corr/PiedeleuKCS15}, and using the same framework also sentence entailment \cite{sadrzadeh2018sentence} and graded hyponymy \cite{bankova2019graded}, since the use of matrices allows the inclusion of correlations between context words. From the computational side, the most recent computational language models \cite{peters-etal-2018-deep,devlin-etal-2019-bert} present contextual embeddings of words as an intrinsic feature. In this paper we aim at reconciling the compositional distributional model and these developments by presenting density matrices as the fundamental representations of words, thus leveraging previous results, and by introducing a refined notion of tensor contraction that can be applied even if we do not assume that we are working with static embeddings coming from the data, thus additionally presenting the possibility of eliminating the distinction between context and target words, because all words can be equally represented with respect to one another. To achieve this, we build the components of the density matrices as covariant or contravariant by introducing a metric that relates them, extending to the interpretation space the notion of directionality of word application, as a direct image of the directional Lambek calculus. After that, we attach permutation operations that act on either type of components to describe derivational ambiguity in a way that keeps multiple readings represented in formally independent vector spaces, thus opening up the possibility of integration between lexical and syntactic ambiguity. 
\\

Section \ref{proofs} introduces our syntactic engine, the Lambek calculus  \textbf{(N)L}$_{/,\bs}$,
together with the Curry-Howard correspondence that associates syntactic derivations
with programs of the ordered lambda calculus $\lambda_{/,\bs}$.
Section \ref{metric} motivates the use of a more refined notion of inner product and introduces the concept of a tensor and tensor contraction as a basis independent application of a dual vector to a vector, and introduces a metric as the mechanism to go from vectors extracted from the data to the dual vectors necessary to perform tensor contraction.
Section \ref{quantum} gives some background on density matrices, and
on ways of capturing the directionality of our syntactic type logic in these semantic spaces using the previously described metric.
Section \ref{semproofs} then turns to the compositional interpretation of the $\lambda_{/,\bs}$ programs
associated with \textbf{(N)L}$_{/,\bs}$ derivations. Section \ref{ambi} shows how the directional density matrix
framework can be used to capture simple forms of derivational ambiguity.

\section{From proofs to programs}\label{proofs}
With his \cite{lambek1958mathematics,lam61} papers, Jim Lambek initiated the `parsing as deduction' method
in computational linguistics: words are assigned formulas of a type logic designed to reason about grammatical
composition; the judgement whether a phrase is well-formed is the outcome of a process of deduction in that
type logic. Lambek's original work was on a calculus of \emph{syntactic} types, which he presented in two versions.
With \textbf{L}$_{/,\bs}$ we refer to the simply typed (implicational) fragment of Lambek's \cite{lambek1958mathematics} associative
syntactic calculus, which assigns types to \emph{strings}; \textbf{NL}$_{/,\bs}$ is the non-associative version
of \cite{lam61}, where types are assigned to \emph{phrases} (bracketed strings)\footnote{Neither of these calculi by itself is
satisfactory for modelling natural language syntax. To handle the well-documented problems of over/undergeneration of
\textbf{(N)L}$_{/,\bs}$ in a principled way, the logics can be extended with modalities that allow for
controlled forms of reordering and/or restructuring. We address these extensions in \cite{correia2020putting}.}.

Van Benthem \cite{vbenthem1983}
added semantics to the equation with his work on \textbf{LP}, a commutative version of the Lambek calculus,
which in retrospect turns out to be a precursor of (multiplicative intuitionistic) linear logic. \textbf{LP}
is a calculus of \emph{semantic} types. Under the Curry-Howard `proofs-as-programs' approach, derivations in \textbf{LP}
are in 1-to-1 correspondence with terms of the (linear) lambda calculus; these terms can be seen as \emph{programs} for
compositional meaning assembly. To establish the connection between syntax and semantics, the Lambek-Van Benthem
framework relies on a homomorphism sending types and proofs of the syntactic calculus to their semantic counterparts.

\begin{figure}
\begin{center}

\[\textrm{Terms:}\qquad\begin{array}{cccccccccccc}
t,u & ::= & x &\mid& \lambda^{r} x.t &\mid& \lambda^{l} x.t &\mid& t\triangleleft u &\mid& u \triangleright t\\
\end{array}\]

Typing rules:

\[\infer[Ax]{\nd{x:A}{x:A}}{}\]
\[\infer[I/]{\nd{\Gamma}{\lambda^{r} x.t:B/A}}{\nd{\Gamma, x:A}{t:B}} 
\qquad
\infer[I\bs]{\nd{\Gamma}{\lambda^{l} x.t:A\bs B}}{\nd{x:A,\Gamma}{t:B}}\]
\[\infer[E/]{\nd{\Gamma,\Delta}{t \triangleleft u:B}}{\nd{\Gamma}{t:B/A} & \nd{\Delta}{u:A}}
\qquad
\infer[E\bs]{\nd{\Gamma,\Delta}{u\triangleright t:B}}{\nd{\Gamma}{u:A} & \nd{\Delta}{t:A\bs B}}\]

\caption{Proofs as programs for \textbf{(N)L}$_{/,\bs}$.}
\label{nlprograms}
\end{center}
\end{figure}

In this paper, rather than defining semantic interpretation on a commutative type logic such as \textbf{LP},
we want to keep the distinction between the left and right implications $\backslash, \slash$ 
of the syntactic calculus in the vector-based semantics we aim for. To achieve this, our programs
for meaning composition use the language of Wansing's \cite{wansing1992} \emph{directional}
lambda calculus $\lambda_{\slash,\backslash}$. Wansing's overall aim is to study how the derivations of a
family of substructural logics can be encoded by typed lambda terms. Formulas, in the substructural setting,
are seen as information pieces, and the proofs manipulating these formulas as
information processing mechanisms, subject to certain conditions that reflect the
presence or absence of structural rules. The terms of $\lambda_{\slash,\backslash}$ faithfully encode
proofs of \textbf{(N)L}$_{/,\bs}$; information pieces, in these logics, cannot be copied or
deleted (absence of Contraction and Weakening), and information processing is sensitive to
the sequential order in which the information pieces are presented (absence of Permutation).

We present the rules of \textbf{(N)L}$_{/,\bs}$ with the
associated terms of $\lambda_{\slash,\backslash}$ in Fig \ref{nlprograms}.
The presentation is in the sequent-style natural deduction format.
The formula language has atomic types (say \emph{s, np, n} for sentences, noun phrases, common nouns)
for complete expressions and implicational types $A\bs B$, $B/A$ for incomplete expressions,
selecting an $A$ argument to the left (resp. right) to form a $B$.

Ignoring the term labeling for a moment,
judgments are of the form $\nd{\Gamma}{A}$, where the antecedent $\Gamma$ is a non-empty list
(for \textbf{L}) or bracketed list (\textbf{NL}) of formulas, and the succedent a single
formula $A$. For each of the type-forming operations, there is an Introduction rule,
and an Elimination rule.

Turning to the Curry-Howard encoding of \textbf{NL}$_{/,\bs}$ proofs, we introduce a language of directional
lambda terms, with variables as atomic expressions, left and right $\lambda$ abstraction, and left and right
application. The inference rules now become \emph{typing} rules for these terms, with judgments of
the form 

\begin{equation}\label{sequent}
    x_1:A_1,\ldots,x_n:A_n\vdash t:B.
\end{equation} The antecedent is a typing environment providing type
declarations for the variables $x_i$; a proof constructs a program $t$ of type $B$ out of these variables.
In the absence of Contraction, Weakening and Permutation structural rules, the program $t$ contains $x_1,\ldots,x_n$
as free variables exactly once, and in that order. Intuitively, one can see a term-labelled proof
as an algorithm to compute a meaning $t$ of type $B$ with parameters $x_i$ of type $A_i$. In parsing
a particular phrase, one substitutes the meaning of the constants (i.e.~words) that
make it up for the parameters of this algorithm.

\section{Directionality in interpretation}\label{metric}

In order to introduce the directionality of the syntactic calculus in the semantic calculus, we expand on the existing literature that uses \textbf{FdVect} as the interpretation category by calling attention to the implied inner product. We introduce a more abstract notion of tensor, tensor contraction and the need to introduce explicitly the existence of a metric, coming from the literature of general relativity, following the treatment in \cite{wald1984general}\footnote{An alternative introductory treatment of tensor calculus can be found in \cite{dullemond1991introduction}.}. Formally, a metric is a function that assigns a distance between two elements of a set, but if applied to the elements of a set that is closed under addition and scalar multiplication, that is, the elements of a vector space, it becomes an inner product. Since we will be looking at vector spaces, we use the terms metric and inner product interchangeably. 

To motivate the need for a more careful treatment regarding the inner product, lets look at a very simple yet illustrative example. Suppose that a certain language model provides word embeddings that correspond to two-dimensional, real valued vectors. In this model, the words "vase" and "wall" have the vector representations $\vec{v}$ and $\vec{w}$, respectively;

\begin{align} \label{1st}
&\Vec{v}=(0,1), \; \vec{w}=(1,0).
\end{align} This representation could mean that they are context words in a count-based model, since they form the standard (orthogonal) basis of $\mathbb{R}^2$, or that they have this particular representation in a particular context-dependent language model. To compute cosine similarity, the notion of Euclidean inner product is used, where the components corresponding to a certain index are multiplied:

\begin{equation}
    \Vec{v}\cdot\vec{w}= 0\cdot1 + 1\cdot 0 = 0,
\end{equation} which we can use to calculate the cosine of the angle $\theta$ between these vectors, 

\begin{equation} \label{cosine}
\cos(\theta)=\frac{\Vec{v}\cdot \vec{w}}{\norm{\Vec{v}}\cdot \norm{\Vec{w}}}= \frac{0\cdot1 + 1\cdot 0 = 0}{1\cdot 1} =0.
\end{equation} Thus, if the representations of these words are orthogonal, then using this measure to evaluate similarity we conclude that these words are not related. However, there is a degree of variation in the vectors that are assigned to the distributional semantics of each word. \textit{Static embeddings} are unique vector representations given by a global analysis of a word over a corpus. The unique vector assigned to the semantics of a word depends on the model used to analyze the data, so different models do not necessarily put out the same vector representations. Alternative to this are \textit{dynamic embeddings}, which assign different vector representations to the same word depending on context, within the same model. 

Therefore, there are at least three ways in which the result of eq.\ref{cosine} and subsequent interpretation can be challenged: 

\begin{enumerate}
    \item \textbf{Static Embeddings.} If the representations come from a count-based model, choosing other words as context words changes the vector representation and therefore these words are not orthogonal to one another anymore; in fact this can happen with any static embedding representation when the basis of the representation changes. Examples of models that give static embeddings are Word2Vec \cite{mikolov2013distributed} and GloVe \cite{pennington2014glove}. \label{1}
    \item \textbf{Dynamic Embeddings.} When the vector representations comes from a context-dependent embedding, changing the context in which the words are evaluated will change their representation, which might not be orthogonal anymore. Dynamic embeddings can be obtained by  i.e. ELMo \cite{peters-etal-2018-deep} and BERT \cite{devlin-etal-2019-bert}.  \label{2}
    \item \textbf{Expectation of meaning.} Human judgements, which are the outcomes of experiments where subjects are explicitly asked to rate the similarity of words, predict that some words should have a degree of relationship. Therefore, the conclusion we derive from orthogonal representations of certain words might not be valid if there is a disagreement with their human assessment. These judgements are condensed in datasets such as the MEN dataset \cite{bruni2014multimodal}.
\end{enumerate} While points \ref{1} and \ref{2} can be related, caution is necessary in establishing that link. On a preliminary inspection, comparing the cosine similarity of context-free embeddings of nouns extracted from pre-trained BERT \cite{devlin-etal-2019-bert} with the normalized human judgements from the MEN dataset \cite{bruni2014multimodal}, we find that the similarity between two words given by the language model is systematically overrated when compared to its human counterpart. One possible explanation is that the language model is comparing all words against one another, so it is an important part of similarity that the two words belong to the the same part of speech, namely nouns, while humans assume that as a condition for similarity evaluation. Further, though we can ask the language model to rate the similarity of words in specific contexts, that has not explicitly been done with human subjects. A more detailed comparison between context-depend representations and human judgement constitutes further research.

One way to reconcile the variability of representations and the notion of similarity is to expand the notion of inner product to be invariant under the change of representations. Suppose now that by points \ref{1} or \ref{2} the representations of "vase" and "wall" change, respectively, to

\begin{align}
&\Vec{v}'=(1,1), \; \vec{w}'=(-1,2).
\end{align} These vectors also form a basis of $\mathbb{R}^2$, but not an orthogonal one. If we use the same measure to compute similarity, taking normalization into account, the Euclidean inner product gives $\Vec{v}'\cdot \vec{w}' =(-1)\cdot 1 + 1\cdot 2 = 1 $ and cosine similarity gives

\begin{equation}\label{cosinesimilarity}
    cos(\theta')=\frac{\Vec{v}'\cdot \vec{w}'}{\norm{\Vec{v}'}\cdot \norm{\Vec{w}'}}= \frac{1}{\sqrt{2}\cdot \sqrt{5}} = \frac{1}{\sqrt{10}}.
\end{equation} If now we have a conflict between which representations are the correct ones, we can look at the human evaluations of similarity. Suppose that it corresponds too to $\frac{1}{\sqrt{10}}$.

We argue in this paper that, by introducing a different notion of inner product, we can fine-tune a relationship between the components of the vectors with the goal to preserve a particular value, for example a human similarity judgement. In this framework, the different representations of words in dynamic embeddings are brought about by a change of basis, similarly to what happens when the context words change in static embeddings, in which case the value of the inner product should be preserved. This can be achieved by describing the inner product as a tensor contraction between a vector and a dual vector, with the latter computed using the metric.

Let $V$ be a finite dimensional vector space and let $V^*$ denote its dual vector space, constituted by the linear maps from $V$ to the field $\mathbb{R}$. A tensor $T$ of type $(k,l)$ over $V$ is a multilinear map

\begin{equation}
  T:  \underbrace{V^* \times \cdots \times V^*}_{k} \times \underbrace{V \times \cdots \times V}_{l} \rightarrow \mathbb{R}.  
\end{equation} Once applied on $k$ dual vectors and $l$ vectors, a tensor outputs an element of the field, in this case a real number. By this token, a tensor of type $(0,1)$ is a dual vector, which is the map from the vector space to the field, and a tensor of type $(1,0)$, being technically the dual of a dual vector, is naturally isomorphic to a vector. Given a basis $E=\{\hat{e}_{i}\}$ in $V$ and its dual basis $\prescript{}{d}{E}=\{\hat{e}^{j}\}$ in $V^*$, with $\hat{e}^{j}(\hat{e}_{i})=\delta_{i}^{j}$, the tensor product between the basis vectors and dual basis vectors forms a basis $B=\{\hat{e}_{i_1} \otimes \cdots \otimes \hat{e}_{i_k} \otimes \hat{e}^{j_1} \otimes \cdots \otimes \hat{e}^{j_l} \}$ of a tensor of type $(k,l)$, allowing the tensor to be expressed with respect to this basis as

\begin{equation} \label{tensor}
T= \sum_{i_1, \ldots, i_k, j_1, \ldots, j_l} T^{i_1\ldots i_k}_{\; \; \; \; \; \; \; \; \; \; \; \; j_1 \ldots j_l} \hat{e}_{i_1} \otimes \cdots \otimes \hat{e}_{i_k} \otimes \hat{e}^{j_1} \otimes \cdots \otimes \hat{e}^{j_l}. 
\end{equation} The basis expansion coefficients $
T^{i_1 \ldots  i_k}_{\; \; \; \; \; \; \; \; \; \;  j_1 \ldots j_l}$ are called the \textit{components} of the tensor.

We can perform two important operations on tensors: apply the tensor product between them, $T'\otimes T$, and contract components of the tensor, $CT$. The first operation happens in the obvious way, while the second corresponds to applying one of the basis dual vectors to a basis vector, resulting in an identification and summing of the corresponding components:

\begin{equation}
(CT)^{i_1\ldots i_{k-1}}_{\; \; \; \; \; \; \; \; \; \; \; \; \; \; \;  j_1 \ldots j_{l-1}}= \sum_\sigma T^{i_1\ldots \sigma \ldots i_{k-1}}_{\; \; \; \;\; \; \; \; \; \;  \; \; \; \; \;\; \; \; \; \; \; j_1 \ldots \sigma \ldots j_{l-1}}.
\end{equation} The outcome is a tensor of type $(k-1, l-1)$. Note that this procedure is basis independent, because of the relationship between the basis and dual basis. For a tensor of type $(1,1)$, which represents a linear operator from $V$ to $V$, tensor contraction corresponds precisely to taking the trace of that operator. To simplify the notation, we will use primed indices instead of numbered ones when the tensors have a low rank.
We define a special $(0,2)$ tensor called a \textit{metric} $d$:

\begin{equation}
    d= \sum_{j, j'} d_{j j'} \hat{e}^{j} \otimes \hat{e}^{j'}.
\end{equation} This tensor is symmetric and non-degenerate. The contraction of this tensor with two vectors $v$ and $w$ gives the value of the inner product: 

\begin{equation}\label{dotprod}
d(v,w)= \sum_{j, j'} d_{j j'} v^{j} w^{j'}.
\end{equation}
 Because of symmetry, $d(v,w)=d(w,v)$, and because of non-degeneracy, the metric is invertible, with its inverse $d^{-1}$ expressed as

\begin{equation}
    d^{-1}= \sum_{i, i'} d^{i i'} \hat{e}_{i} \otimes \hat{e}_{i'}.
\end{equation}  

Given that the elements extracted from the data are elements of $V$, the contractions that need to be performed, for example for the application of the compositionality principle in vector spaces, must involve a passage from vectors to dual vectors as seen in the DisCoCat model, before contraction takes place. The metric can be used to define a canonical map between $V$ and $V^*$ via the partial map that is obtained when only one vector is used as an argument of the metric, giving rise to the dual vector $\prescript{}{d}{v}: v \mapsto d(-,v)$, with the slash indicating the empty argument slot:

\begin{equation}\label{innerprod}
    d(v,w)\equiv d(v,-)(w) \equiv \prescript{}{d}{v(w)}.
\end{equation} This formulation is basis independent, since it results from tensor contraction. Once a basis is defined, the resulting dual vector can be expressed as 

\begin{equation}
  v^d = \sum_{i, j, j'} d_{jj'} v^{i} \hat{e}^{j} \otimes \hat{e}^{j'} (\hat{e}_{i}) = \sum_{ j, j'} d_{jj'} v^{j'} \hat{e}^{j} = \sum_{j'} v_{j'}  \hat{e}^{j'},  
\end{equation}where we rewrite $v_{j'}=\sum_{j} d_{j j'} v^{j'}$.

We call the components of vectors, with indices "up", the \textit{contravariant} components, and those of dual vectors, with indices "down", the \textit{covariant} components. Thus, consistent with our notation, the metric can be used to "lower" or "raise" indices, applying contraction between the metric and the tensor and relabeling the components:

\begin{align}
    d(T)=&\sum_{i_1,\ldots ,i_k,j_1, \ldots, j_{l+2}} d_{j_{l+1},j_{l+2}} T^{i_1,\ldots,i_k}_{\; \; \; \; \; \; \; \; \; \; \; \; \; \;  j_1, \ldots, j_l}  \hat{e}^{j_{l+1}} \otimes \hat{e}^{j_{l+2}} (\hat{e}_{i_1}) \otimes \ldots \otimes \hat{e}^{j_l} \nonumber \\
&= \sum_{i_1,\ldots ,i_k,j_1, \ldots, j_{l+1}} d_{j_{l+1},i_1} T^{i_1,\ldots ,i_k}_{\; \; \; \; \; \; \; \; \; \; \; \; \; \;  j_1, \ldots, j_l} \hat{e}^{j_{l+1}} \otimes \hat{e}_{i_2} \otimes \ldots \otimes \hat{e}^{j_l} \nonumber \\
& =   \sum_{i_2,\ldots ,i_k,j_1, \ldots, j_{l+1}} \tensor{T}{_{j_{l+1}}^{\; \;\; \; \; i_2,\ldots ,i_k}_{ j_1, \ldots, j_l}} \hat{e}^{j_{l+1}} \otimes \hat{e}_{i_2} \otimes \ldots \otimes \hat{e}^{j_l}.
\end{align}

The effect of the metric on a tensor can be captured by seeing how we rewrite the components of some example tensors:

\begin{itemize}
    \item $\sum_{j'} d_{j j'} \tensor{T}{^{j'}_{\; \; j'' }} = T_{j j''};$
    \item $\sum_{i'} \tensor{T}{^{i}_{\; \; i'}} d^{i' i''} = T^{i i''};$
    \item $\sum_{j', j'''} d_{j j'} d_{j'' j'''} \tensor{T}{^{j' j'''}} = \tensor{T}{_{j j''}}. $
\end{itemize} Most importantly, a proper tensor is only defined in the form of eq.\ref{tensor}, so whenever we have a tensor that has components "up" and "down" in different orders, for example in $\tensor{T}{_{j}^{\;i}}$, this is in fact a tensor of type $(1,1)$ of which the actual value of the components is 

\begin{equation} \label{tensorhigher}
  \sum_{i', j'}  d^{i i'} d_{j j'} \tensor{T}{^{j'}_{\;i'}}.  
\end{equation}
\\

Returning to our toy example with the words "vase" and "wall", we can look at the change in vector representations as a change of basis $\hat{e}_i= \sum_{i'} \Lambda_i^{\; i'} \hat{e'}_{i'}$:

\begin{equation}
\vec{v}=\sum_i v^i \hat{e}_i= \sum_{ii'} v^i \Lambda^{\;i'}_ { i} \hat{e'}_{i'} = \sum_{i'} v'^{i'} \hat{e'}_{i'},
\end{equation} corresponding to a change in the vector components $v'^{i'}=v^i \Lambda^{\;i'}_ { i}$. The components of the metric also change with the basis: \begin{equation} \label{change}
    d'_{j''j'''}=  \Lambda_{j'''}^{\; \; j'} \Lambda_{j''}^{\; \; j} d_{jj'}. 
\end{equation} With this change, we can show that inner product remains invariant under a basis change:

\begin{equation}
   w'^{i'}v'_{i'} = w'^{i'}  v'^{j'} d'_{j' i'} =   w'^{i'}  v'^{j'}  \Lambda_{i'}^{\; \; i} \Lambda_{j'}^{\; \; j} d_{ji}  =  w^i v^j d_{ji} = w^i v_i.
\end{equation} In this way, finding the right metric allow us to preserve a value that is constant in the face of context dependent representations. Assuming a metric that has the following matrix representation in the standard basis,

\begin{equation}
   d=\begin{pmatrix}
2 & 1\\ 
1 & 5
\end{pmatrix},
\end{equation} its application to the vector elements in eqs.\ref{1st} gives a value of the inner product calculated in the new representation:

\begin{equation}
    v'_{i'} w'^{i'}= \begin{pmatrix}
1 & 0
\end{pmatrix} \begin{pmatrix}
2 & 1\\ 
1 & 5
\end{pmatrix} \begin{pmatrix}
0\\
1
\end{pmatrix}=1.
\end{equation} Since the norm of the vectors has to be calculated using the same notion of inner product,  

\begin{equation}
    \norm{\vec{v}}= \sqrt{v^i g_{ij} v^j},
\end{equation} we find exactly the cosine similarity calculated in eq. \ref{cosinesimilarity}. Note that this formalism allows us to deal with non-orthogonal basis, but does not require it: in fact, there is an implicit metric already when we compute the Euclidean inner product in eq.\ref{1st}, given by $d_{orth}= \begin{pmatrix}
1 & 0\\ 
0 & 1
\end{pmatrix}$ in the standard basis, which should be used in the case of an orthonormal basis.

Since these new tools allow us to preserve a quantity in the face of a change of representation, we can start reversing the question on similarity: given a certain human judgement on similarity, or another constant of interest, what is the metric that preserves it across different representations \footnote{In case the quantity we wish to preserve is other than that of the Euclidean inner product in either representation, there is an option to expand the vector representation of our words by adding vector components that act as parameters, to ensure that the quantity is indeed conserved. This would be similar to the role played by the time dimension in Einstein's relativity theory.}? Once the vector spaces are endowed with specific metrics, the new inner product definitions permeate all higher-rank tensor contractions that are performed between higher and lower rank tensors, namely the ones that will be used in the interpretation of the Lambek rules\footnote{Using this formalism, we can replace the unit and counit maps $\epsilon$ and $\eta$ maps of the compact closed category \textbf{FdVect} by
$$\eta ^l :\mathbb{R} \rightarrow V \otimes V^* :: 1 \mapsto \mathbb{1} \otimes d(\mathbb{1},-) $$
$$\eta ^r :\mathbb{R} \rightarrow V^* \otimes V :: 1 \mapsto d(-,\mathbb{1}) \otimes \mathbb{1} $$
$$\epsilon ^l : V^* \otimes V \rightarrow \mathbb{R} :: d(-,v)\otimes u \mapsto d(u,v) $$
$$\epsilon ^r : V \otimes V^* \rightarrow \mathbb{R} :: v \otimes d(u,-) \mapsto d(u,v).$$}, and can further be extended to density matrices.


\subsection{Metric in Dirac Notation}

We want to lift our description to the realm of density matrices. We now show how the concept of metric can also be introduced in that description, such that the previously described advantages carry over.

Dirac notation is the usual notation for vectors in the quantum mechanics literature. To make the bridge with the previous concepts from tensor calculus, we introduce it simply as a different way to represent the basis and dual basis of a vector space. Let us rename their elements as \textit{kets} $\ket{i} \equiv \hat{e}_{i} $ and as \textit{bras} $\bra{j} \equiv \hat{e}^{j}$. The fact that the bases are dual to one another is expressed by the orthogonality condition $\braket{j}{i}=\delta_{ij}$, which, if the vector basis elements are orthogonal to each other, is equivalent to applying the Euclidean metric to $\ket{i}$ and $\ket{j}$. Using Dirac notation,  a vector and dual vector are represented as $ v\equiv \ket{v}= \sum_i v^i \ket{i} $ and $v^d \equiv \bra{u} = \sum_j v_j \bra{j}$ \footnote{For orthonormal basis over the field of complex numbers, the covariant components are simply given by the complex conjugate of the contravariant ones, $v_i=\bar{v}^i$.}.
If the basis elements are not orthogonal, this mapping has to be done through a more involved metric. To express this, in this paper we introduce a modified Dirac notation over the field of real numbers, inspired by the one used in \cite{georgi1982lie} for the treatment of quantum states related by a specific group structure \footnote{This treatment can be extended to the field of complex numbers by considering that the metric has conjugate symmetry, $d_{ij}=\bar{d}_{ji}$ \cite{sadun2007applied}.}. The previous basis elements of $V$ are written now as $\ket{_i}\equiv \hat{e}_{i}$ and the corresponding dual basis as $\bra{^j} \equiv \hat{e}^{j}$, such that $\braket{^j}{_i}=\delta_i^j$.
In this basis, the metric is expanded as $d=\sum_{j,j'} d_{j' j} \bra{^{j}}\otimes \bra{^{j'}}$ while the inverse metric is expressed as $d^{-1}=\sum_{i i'} d^{i'i} \ket{_{i}} \otimes \ket{_{i'}}$. The elements of the metric and inverse metric are related by $\sum_{i} d_{j'i'}d^{i' i}=\delta_{j'}^{i'}$. Applying the metric to a basis element of $V$, we get

\begin{equation}\label{bra2}
    \bra{_i} \equiv d(-,\ket{_i})=\sum_{j j'} d_{j'j}  \bra{^{j}}\otimes \braket{^{j'}}{_i} = \sum_{j} d_{i j} \bra{^{j}}. 
\end{equation} Acting with this on $\ket{_{i'}}$ to extract the value of the inner product, the following formulations are equivalent:

\begin{align} 
  d(\ket{_{i'}},\ket{_i})= d(-,\ket{_i})\ket{_{i'}} =  \sum_{j} d_{ij} \braket{^{j}}{_{i'}} = \braket{_i}{_{i'}} = d_{ii'}. 
\end{align} When the inverse metric is applied to $\bra{^j}$ it gives

\begin{equation}\label{ket2}
    \ket{^j} \equiv d\left(-,\bra{^j}\right)= \bra{^{j}} \sum_{i i'} d^{i'i} \ket{_{i}} \otimes \ket{_{i'}} = \sum_{i'} d^{i'j } \ket{_{i'}},
\end{equation} with a subsequent application on $\bra{^{j'}}$ giving

\begin{equation}
    d^{-1}\left(\bra{^{j'}},\bra{^j}\right) = \bra{^{j'}} d\left(-,\bra{^j} \right) = \bra{^{j'}} \sum_{i'} d^{i'j} \ket{_{i'}} = \braket{^{j'}}{^j} = d^{j'j}.
\end{equation} Consistently, we can calculate the value of the new bras and kets defined in eqs.\ref{bra2} and \ref{ket2} applied to one other, showing that they too form a basis/dual basis pair:

\begin{equation}
\braket{_i}{^j}= \sum_{j'} d_{ ij'} \bra{^{j'}} \sum_{i'} d^{ i'j} \ket{_{i'}}  = \sum_{i' j'}  d_{ij'}  d^{i'j} \braket{^{j'}}{_{i'}} = \sum_{j'}d_{ij'}  d^{j' j}=\delta_i^j.    
\end{equation}If the basis elements are orthogonal, the components of the metric and inverse metric coincide with the orthogonality condition.

\section{Density Matrices: Capturing Directionality}\label{quantum}
The semantic spaces we envisage for the interpretation of the syntactic calculus are
density matrices.
A \textit{density matrix} or density operator is used in quantum mechanics to describe systems for which the state is not completely known.
For lexical semantics, it can be used to describe the meaning of a word by placing distributional information on its components.
As standardly presented \footnote{A background for the non-physics reader can be found in \cite{nielsen2002quantum}.},
density matrices that are defined on a tensor product space indicate no preference with respect to contraction from the left or from the right.
Because we want to keep the distinction between left and right implications in the semantics, we set up the interpretation of
composite spaces in such a way that they indicate which parts will and will not contract with other density matrices.

The \emph{basic} building blocks of the interpretation are density matrix spaces $\tilde{V} \equiv V \otimes V^*$. For this composite space, we choose the basis formed by $\ket{_i}$ tensored with $\bra{_{i'}}$, $\tilde{E}=\left\{ \ket{_i}\bra{_{i'}} \right\} = \left\{\tilde{E}_J\right\}$. Carrying over the notion of duality to the density matrix space, we define the dual density matrix space $\tilde{V}^* \equiv V \otimes V^*$. The dual basis in this space is the map that takes each basis element of $\tilde{V}$ and returns the appropriate orthogonality conditions. It is formed by $\bra{^j}$ tensored with $\ket{^{j'}}$, $\prescript{}{d}{\tilde{E}}=\left\{ \ket{^{j'}}\bra{^j} \right\} = \left\{\tilde{E}^J\right\}$ , and is applied on the basis vectors of $\tilde{V}$ via the trace operation 

\begin{align}
& \tilde{E}^{J}\left(\tilde{E}_I \right)= \Tr \left(  \ket{_i}\braket{_{i'}}{^{j'}}\bra{^j}  \right) = \sum_l \braket{^l}{_i}\braket{_{i'}}{^{j'}}\braket{^j}{_l} \nonumber \\
& = \sum_{jj'} \braket{^j}{_i} \braket{_{j'}}{^{i'}} \delta_i^j \delta_{i'}^{j'} \equiv \delta_{I}^{J}.
\end{align}

Because density operators are hermitian, their matrices do not change under conjugate transposition, which extends to elements of the basis of the density matrix space. In this way, we can extend our notion of metric to the space of density matrices, where a new metric $D$ emerges from $d$, expanded in the basis of $V^*$ as

\begin{align}
   &  D=\sum_{J, J'} D_{J J'} \tilde{E}^{J}\otimes \tilde{E}^{J'} \label{bigmetric1} \\ 
   &=\sum_{j j', j'' j'''} d_{j'' j'} d_{j''' j} \ket{^{j'}}\bra{^{j}} \otimes \ket{^{j'''}}\bra{^{j''}} \label{bigmetric2}.
\end{align}

We can see how both definitions are equivalent by their action on a density matrix tensor $ T \equiv \sum_I T^I \tilde{E}_I \equiv \sum_{ii'} T^{ii'} \ket{_i}\bra{_{i'}}$.
Staying at the level of $\tilde{V}$ and $\tilde{V}^*$, we use eq.\ref{bigmetric1} to  obtain

\begin{align}
   & D(-, T) =  \sum_{I,J, J'} D_{J J'} T^I \tilde{E}^{J}\otimes \tilde{E}^{J'} \left( \tilde{E}_I \right) = \sum_{I,J, J'}  D_{J J'} T^I \tilde{E}^{J} \delta_I^{J'} \nonumber \\
  &  =  \sum_{J, J'} D_{J J'}  T^{J'} \tilde{E}^{J} \equiv \sum_{J} T_{J} \tilde{E}^{J} = \sum_{jj'} T_{j'j} \ket{^{j'}}\bra{^{j}}, 
\end{align} where we redefine $T_{J} \equiv D_{J J'}  T^{J'}$, thus establishing covariance and contravariance of the tensor components defined over the density matrix space. Looking in its turn at the level of $V$ and $V^*$, using eq.\ref{bigmetric2}, we see that both definitions are equivalent:

\begin{align}
&  D(-, T) = \sum_{ii',jj',j'' j'''} T^{ii'}d_{j'' j'} d_{j''' j} \ket{^{j'}}\bra{^{j}}  \otimes \Tr \left( \ket{^{j'''}}\braket{^{j''}}{_i}\bra{_{i'}} \right) \nonumber \\
& = \sum_{ii',jj',j'' j'''} T^{ii'}d_{j'' j'} d_{j'''j} \delta^{j''}_{i} \delta^{j'''}_{i'} \ket{^{j'}}\bra{^{j}} \nonumber\\
&= \sum_{ii' j j'} T^{ii'} d_{i j'} d_{i'j} \ket{^{j'}} \bra{^j} \equiv \sum_{jj'} T_{jj'} \ket{^{j'}} \bra{^{j}},
\end{align} where we rewrite $T_{jj'} \equiv T^{ii'} d_{i j'} d_{i'j}$  \footnote{Here we can compare our formalism to that of the compact closed category of completely positive maps \textbf{CPM(FdVect)} developed in \cite{selinger2007dagger}. The categorical treatment applies here at a higher level, however, introducing the metric defines explicitely the canonical isomorphisms $V \cong V^*$ and $\tilde{V} \cong \tilde{V}^*$, which trickles down to knowing exactly how the symmetry of the tensor product acts on the compenents of a tensor: $ \sigma_{V,V^*}: V^* \otimes V \rightarrow V \otimes V^* :: \sum_{ij} \tensor{T}{_{i}^{\;j}} \hat{e}^i \otimes \hat{e}_j \mapsto \sum_{ii', jj'}  d^{i i'} d_{j j'} \tensor{T}{^{j'}_{\;i'}} \hat{e}_{i} \otimes \hat{e}^{j}$.}.

From these basic building blocks,
\emph{composite} spaces are formed via the binary operation $\otimes$ (tensor product) and a unary operation $()^*$ (dual functor) 
that sends the elements of a density matrix basis to its dual basis, using the metric defined above. In the notation, we use $\tilde{A}$ for density matrix spaces (basic or compound),
and $\rho$, or subscripted $\rho_x, \rho_y, \rho_z,\ldots \in \tilde{A}$ for elements of such spaces. 
The $()^*$ operation is involutive; it interacts with the tensor product as $(\tilde{A} \otimes \tilde{B})^*= \tilde{B}^* \otimes \tilde{A}^*$  and acts as identity on matrix multiplication.

Below in (\dag) is the general form of a density matrix defined on a single space in the standard basis,
and (\ddag) in the dual basis:

\[(\dag)\quad \rho_x^{\tilde{A}}= \sum_{ii'} X^{ii'} \ket{_i} \prescript{}{\tilde{A}}{\bra{_{i'}}},
\qquad
(\ddag)\quad \rho_x^{\tilde{A}^*}= \sum_{jj'} X_{j'j} \ket{^{j'}} \prescript{}{\tilde{A}^*}{\bra{^{j}}}.
\] Over the density matrix spaces, we can see these matrices as \textit{tensors} as we defined them previously, with $X^I \equiv X^{ii'}$ the \textit{contravariant} components and with $X_{J'} \equiv X_{j'j}$ the \textit{covariant} components. 

A density matrix of a composite space can be an element of the tensor product space between the standard space and the dual space either from the left or from the right:

\begin{equation}
 \rho_y^{{\tilde{A}} \otimes {\tilde{B}}^{*}} = \sum_{ii',jj'} Y^{ii'}_{\; \; j'j} \ket{_{ i}^{\; \; j'}} \prescript{}{\tilde{A} \otimes \tilde{B}^*}{\bra{_{i'}^{\; \; j}}};  
\end{equation}

\begin{equation}
  \rho_w^{\tilde{B}^*  \otimes \tilde{A}} = \sum_{ii',jj'} W_{j'j}^{\; \; ii'} \ket{^{j'}_{\; \;  i}} \prescript{}{\tilde{B}^* \otimes \tilde{A}}{\bra{^{j}_{\; \; i'}}}.  
\end{equation} Although both tensors are of the form $(1,1)$, the last one is a tensor with components $Y_{\; J'}^{I}$, which relate with a true tensor form by $D^{II'} Y^{\; J}_{I'} D_{JJ'}$.
Recursively, density matrices that live in higher-rank tensor product spaces can be constructed, taking a tensor product with the dual basis either from the left or from the right. Multiplication between two density matrices of a standard and a dual space follows the rules of tensor contraction:

\begin{align}
&\rho_y^{\tilde{A}^*} \cdot \rho_x^{\tilde{A}} =  \sum_{jj'} Y_{j'j} \ket{^{j'}} \prescript{}{\tilde{A}^*}{\bra{^{j}}} \cdot \sum_{ii'} \prescript{}{}{X^{ii'}} \ket{_i} \prescript{}{\tilde{A}}{\bra{_{i'}}}  = \sum_{i',jj'} Y_{j'j} X^{ji'} \ket{^{j'}} \prescript{}{\tilde{A}}{\bra{_{i'}}}. 
\end{align}

\begin{align}
&  \rho_x^{\tilde{A}} \cdot \rho_y^{\tilde{A}^*} = \sum_{ii'} \prescript{}{}{X^{ii'}} \ket{_i}  \prescript{}{\tilde{A}}{\bra{_{i'}}} \cdot \sum_{jj'} Y_{j'j} \ket{^{j'}} \prescript{}{\tilde{A}^*}{\bra{^{j}}}  = \sum_{i,jj'} X^{ij'} Y_{j'j}  \ket{_i} \prescript{}{\tilde{A}}{\bra{^{j}}},
\end{align} respecting the directionality of composition. To achieve full contraction, the trace in the appropriate space is applied, corresponding to a partial trace if the tensors involve more spaces:

\begin{align}
  \Tr_{\tilde{A}} \left( \sum_{i',jj'} Y_{j'j} X^{ji'} \ket{^{j'}} \prescript{}{\tilde{A}}{\bra{_{i'}}}  \right) =  \sum_{l,i',jj'} Y_{j'j} X^{ji'} \prescript{}{\tilde{A}}{\braket{_{l}}{^{j'}}_{\tilde{A}^*}} \prescript{}{\tilde{A}}{\braket{_{i'}}{^{l}}_{\tilde{A}^*}} = \sum_{jj'} Y_{j'j} X^{jj'},  
\end{align}

\begin{align}
\Tr_{\tilde{A}} \left(  \sum_{i,jj'} X^{ij'} Y_{j'j}  \ket{^i} \prescript{}{\tilde{A}}{\bra{_{j}}} \right) = \sum_{l,j',ij} X^{ij'} Y_{j'j} \prescript{}{\tilde{A}^{*}}{\braket{_l}{^i}_{\tilde{A}}} \prescript{}{\tilde{A}^*}{\braket{_j}{^l}_{\tilde{A}}} = \sum_{jj'} X^{jj'} Y_{j'j}.
\end{align} We see that the cyclic property of the trace is preserved. 

In \S\ref{ambi} we will be dealing with derivational ambiguity, and for that the concepts of \textit{subsystem} and \textit{permutation operation} introduced here will be useful. 
A subsystem can be thought of as a copy of a space, described using the same basis, but formally treated as a different space. In practice, this means that different subsystems do not interact with one another. In the quantum setting, they represent independent identical quantum systems. For example, when we want to describe the spin states of two electrons, despite the fact that each spin state is defined on the same basis, it is necessary to distinguish which electron is in which state and so each is attributed to their own subsystem. Starting from a space $\tilde{A}$, two different subsystems are referred to as $\tilde{A_1}$ and $\tilde{A_2}$. If different words are described in the same space, subsystems can be used to formally assign them to different spaces. The permutation operation extends naturally from the one in standard quantum mechanics. We define two permutation operators: $P^{\tilde{A_1} \tilde{A_2}}$ permutes the elements of the basis of the respective spaces, while $P_{\tilde{A_1} \tilde{A_2}}$ permutes the elements of the dual basis. 
If only one set of basis elements is inside the scope of the permutation operators, then either the subsystem assignment changes,

\begin{equation}\label{permone}
 P^{\tilde{A_1} \tilde{A_2}}  \ket{_i} \prescript{}{\tilde{A_1}}{\bra{_{i'}}} P^{\tilde{A_1} \tilde{A_2}} = \ket{_i} \prescript{}{\tilde{A_2}}{\bra{_{i'}}}; \qquad  P_{\tilde{A_1} \tilde{A_2}}  \ket{^{i'}} \prescript{}{\tilde{A_1}^*}{\bra{^{i}}} P_{\tilde{A_1} \tilde{A_2}} = \ket{^{i'}} \prescript{}{\tilde{A_2}^*}{\bra{^{i}}}; 
\end{equation} or the respective space of tracing changes,

\begin{equation}\label{permtrace}
\Tr_{\tilde{A_1}} \left( P_{\tilde{A_1} \tilde{A_2}} \ket{_{i'}} \prescript{}{\tilde{A_2}^*}{\bra{_{i}}} P_{\tilde{A_1} \tilde{A_2}} \right) =  \Tr_{\tilde{A_2}} \left( \ket{_{i'}} \prescript{}{\tilde{A_2}^*}{\bra{_{i}}}  \right).   
\end{equation} Note that permutations take precedence over traces. If two words are assigned to different subsystems, the permutations act to swap their space assignment\footnote{We define this as a shorthand application of the permutation operations as defined in eq.\ref{permone}, such that eq.\ref{permdown} can be calculated w.r.t. that definition as $$ P^{\tilde{A_1} \tilde{A_2}}  \ket{_i}_{\tilde{A_1}} \left(\prescript{}{\tilde{A_1}}{\bra{_{i'}}}  P^{\tilde{A_1} \tilde{A_2}} \right) \otimes \left( P^{\tilde{A_1} \tilde{A_2}} \ket{_j}_{\tilde{A_2}} \right) \prescript{}{\tilde{A_2}}{\bra{_{j'}}} P^{\tilde{A_1} \tilde{A_2}}$$ $$ = P^{\tilde{A_1} \tilde{A_2}}  \ket{_i}_{\tilde{A_1}} \prescript{}{\tilde{A_2}}{\bra{_{i'}}}  \otimes  \ket{_j}_{\tilde{A_1}} \prescript{}{\tilde{A_2}}{\bra{_{j'}}} P^{\tilde{A_1} \tilde{A_2}} = \ket{_i} \prescript{}{\tilde{A_2}}{\bra{_{i'}}} \otimes\ket{_j} \prescript{}{\tilde{A_1}}{\bra{_{j'}}},$$ and similarly for eq.\ref{permup}.}:

\begin{equation}\label{permdown}
 P^{\tilde{A_1} \tilde{A_2}}  \ket{_i} \prescript{}{\tilde{A_1}}{\bra{_{i'}}} \otimes\ket{_j} \prescript{}{\tilde{A_2}}{\bra{_{j'}}} P^{\tilde{A_1} \tilde{A_2}} =  \ket{_i} \prescript{}{\tilde{A_2}}{\bra{_{i'}}} \otimes\ket{_j} \prescript{}{\tilde{A_1}}{\bra{_{j'}}},    
\end{equation}

\begin{equation}\label{permup}
   P_{\tilde{A_1} \tilde{A_2}} \ket{^{i'}} \prescript{}{\tilde{A_1}^*}{\bra{^{i}}} \otimes\ket{^{j'}} \prescript{}{\tilde{A_2}^*}{\bra{^{j}}}  P_{\tilde{A_1} \tilde{A_2}} =  \ket{^{i'}} \prescript{}{\tilde{A_2}^*}{\bra{^{i}}} \otimes\ket{^{j'}} \prescript{}{\tilde{A_1}^*}{\bra{^{j}}}. 
\end{equation}   If no word has that subsystem assignment then the permutation has no effect.

\section{Interpreting Lambek Calculus derivations}\label{semproofs}
Let us turn now to the syntax-semantics interface, which takes the form of a homomorphism
sending the types and derivations of the syntactic front end \textbf{(N)L}$_{\slash,\backslash}$ to their semantic counterparts.
Consider first the action of the interpretation homomorphism on \emph{types}. We write 
$\lceil . \rceil$ for the map that sends syntactic types to the interpreting semantic spaces.
For primitive types we set
\begin{equation}
     \lceil s \rceil = \tilde{S},\;  \lceil np \rceil = \lceil n \rceil = \tilde{N},
\end{equation}
with $S$ the vector space for sentence meanings and $N$ the space for nominal expressions (common nouns, full noun phrases).
For compound types we have

\begin{equation}
    \lceil A/B \rceil= \lceil A \rceil \otimes \lceil B \rceil ^* , \; \text{and} \;  \lceil A \backslash B \rceil=  \lceil A\rceil^*  \otimes \lceil B\rceil .
\end{equation}
Given semantic spaces for the syntactic types, we can turn to the interpretation of the syntactic \emph{derivations}, as coded by
their $\lambda_{\slash,\bs}$ proof terms. We write $\Dg{\cdot}$ for the map that associates each term $t$ of type $A$ with
a semantic value, i.e.~an element of $\Spc{A}$, the semantic space where meanings of type $A$ live.
The map $\llbracket . \rrbracket$ is defined relative to a assignment function $g$
that provides a semantic value for
the basic building blocks, viz.~the variables that label the axiom leaves of a proof. 
As we saw above, a proof term is a generic meaning recipe that abstracts from particular
lexical meanings. Specific lexical items, as we will see in \S\ref{ambi}, have the status of \emph{constants}.
These constants are mapped to their distributional meaning by an interpretation function $I$. The distributional meaning corresponds to the embeddings assigned by a particular model to the lexicon.
Below we show that this calculus is sound with respect to the semantics of section \ref{quantum}.


\paragraph*{Axiom}
\begin{equation}
  \left\llbracket x^A \right\rrbracket_g = g(x^A) = \prescript{}{}{\rho_x^{\lceil A \rceil}} = \sum_{ii'} \prescript{}{}{X^{ii'}} \ket{_i} \prescript{}{\lceil A \rceil}{\bra{_{i'}}}.  
\end{equation}

\paragraph{Elimination} Recall the inference rules of Fig \ref{nlprograms}.

\noindent
$E_\slash$: Premises $t^{B \slash A}$, $u^A$; conclusion $(t\triangleleft u)^{B}$:

\begin{align}
& 
\left\llbracket (t\triangleleft u)^B \right\rrbracket_g   \equiv \Tr_{\lceil A \rceil} \left(\left\llbracket t^{B/A} \right\rrbracket_g \cdot  \left\llbracket u^A \right\rrbracket_g \right) \\
& = \Tr_{\lceil A \rceil} \left( \sum_{ii',jj'} \prescript{}{}{T^{ii'}_{\; \; j'j}} \ket{_{i}^{\;j'}} \prescript{}{\lceil B \rceil \otimes \lceil A \rceil^*}{\bra{_{i'}^{\;j}}} \cdot \sum_{kk'} \prescript{}{}{U^{kk'}} \ket{_{k}} \prescript{}{\lceil A \rceil}{\bra{_{k'}}} \right)\\
&=\sum_{ii',jj'} \sum_{kk'} \prescript{}{}{T^{ii'}_{\; \; j'j}} \cdot \prescript{}{}{U^{kk'}} \, \delta^{j}_k \delta^{j'}_{k'}     \ket{_{i}} \prescript{}{\lceil B \rceil}{\bra{_{i'}}} = \sum_{ii',jj'} \prescript{}{}{T^{ii'}_{\; \; j'j}} \cdot \prescript{}{}{U^{jj'}} \ket{_{i}} \prescript{}{\lceil B \rceil}{\bra{_{i'}}}.
\end{align} 

\noindent
$E_\bs$: Premises $u^A$, $t^{A\bs B}$; conclusion $(u\triangleright t)^{B}$:

\begin{align}
& 
\left\llbracket (u\triangleright t)^B \right\rrbracket_g \equiv \Tr_{\lceil A \rceil} \left( \left\llbracket u^A \right\rrbracket_g \cdot \left\llbracket t^{A\backslash B} \right\rrbracket_g  \right) \\
& = \Tr_{\lceil A \rceil} \left(  \sum_{kk'}  \prescript{}{}{U^{kk'}} \ket{_{k}} \prescript{}{\lceil A \rceil}{\bra{_{k}'}}\cdot  \sum_{ii',jj'}   \prescript{}{}{T^{\; \; ii'}_{jj}}  \ket{^{j'}_{\;i}} \prescript{}{\lceil A \rceil^* \otimes \lceil B \rceil}{\bra{^{j}_{\;i'}}} \right)= \\
&= \sum_{kk'}  \sum_{ii',jj'} \prescript{}{}{U^{kk'}} \cdot \prescript{}{}{T^{\; \;ii'}_{j'j}} \,  \delta^{j}_k \delta^{j'}_{k'}  \ket{_{i}} \prescript{}{\lceil B \rceil}{\bra{_{i'}}} = \sum_{ii',jj'}   \prescript{}{}{U^{jj'}} \cdot \prescript{}{}{T^{\; \; \; ii'}_{j'j}} \ket{_{i}} \prescript{}{\lceil B \rceil}{\bra{_{i'}}}.
\end{align} 

\paragraph{Introduction}

\noindent
$I_\slash$: Premise $t^{B}$, with $x^{A}$ as its rightmost parameter; conclusion $( \lambda^r x. t )^{B\slash A}$:

\begin{align}
& 
\left\llbracket \left( \lambda^r x. t \right)^{B\slash A} \right\rrbracket_{g} \equiv \sum_{kk'} \left(\llbracket t^B \rrbracket_{g^{x}_{kk'}} \otimes \ket{^{k'}_{}} \prescript{}{\lceil A \rceil^*}{\bra{^{k}}} \right)
\end{align}

\noindent
$I_\bs$: Premise $t^{B}$, with $x^{A}$ as its leftmost parameter; conclusion $( \lambda^l x. t )^{A\bs B}$:

\begin{align}
& 
\left\llbracket \left( \lambda^l x. t \right)^{A\bs B} \right\rrbracket_{g} \equiv \sum_{kk'} \left( \ket{^{k'}_{}} \prescript{}{\lceil A \rceil^*}{\bra{^{k}}} \otimes \llbracket t^B \rrbracket_{g^{x}_{kk'}}   \right)
\end{align}
Here $g^{x}_{kk'}$ is the assignment exactly like $g$ except possibly for the parametric variable $x$ which takes the value
of the basis element $\ket{_{k}} \prescript{}{\lceil A \rceil}{\bra{_{k'}}}$.
More generally, the interpretation of the introduction rules lives in a compound density matrix space representing a
linear map from $\tilde{A}$ to $\tilde{B}$. The semantic value of that map, applied to any object $m\in\tilde{A}$, is given by
$\llbracket t^B \rrbracket_{g'}$, where $g'$ is the assignment exactly like $g$ except possibly for the bound variable $x^A$, which is
assigned the value $m$. 
Note that now, given the introduction of the metric, the interpretations of $A\slash B$ and $B \backslash A$ are related by it: if the components of the first are $\tensor{T}{_{J}^{\,I}}$, then those of the second are given by those in eq.\ref{tensorhigher} adapted for density matrices. This is what introduces directionality in our interpretation: using the metric, we can extract a certain representation for a function word and distinguish by the values of the components whether it will contract from the left or from the right.

\section{Derivational Ambiguity}\label{ambi}

The density matrix construction has can be successfully used to address lexical ambiguity \cite{piedeleu2014ambiguity}, as well as lexical and sentence entailment \cite{sadrzadeh2018sentence,bankova2019graded}, where different measures of entropy are used to perform the disambiguation. Here we arrive at disambiguation in a different way, by storing in the diagonal elements of a higher order density matrix the different interpretations that result from the different contractions that the proof-as-programs prescribes. This is possible due to the the set-up that is formed by a multi-partite density matrices space, so that, by making use of permutation operations, it happens automatically that the two meanings are expressed independently. This is useful because it can be integrated with a lexical interpretation in density matrices optimized to other tasks, such as lexical ambiguity or entailment. It is also appropriate to treat the existence of these ambiguities in the context of incrementality, since it keeps the meanings separated in their interaction with posterior fragments. 

We give a simple example of how the trace machinery can be used on an ambiguous fragment, providing a passage from one reading to the other at the interpretation level, and how the descriptions are kept separated. For this application, the coefficients in the interpretation of the words contain distributional information harvested from data, either from a count-base model or a more sophisticated language model. The final coefficient of each outcomes is the vector-based representation of that reading.

We illustrate the construction with the phrase "tall person from Spain".
The lexicon below has the syntactic type assignments and the corresponding semantic spaces.

\[\begin{array}{r|c|l}
 & \textrm{syn type $A$} & \lceil A\rceil\\\hline
\textrm{tall} & n/n & N^*\otimes N \otimes ( N^*\otimes N )^*\\
\textrm{person} & n & N^*\otimes N\\
\textrm{from} & (n\bs n)/np & \left(N^*\otimes N \right)^* \otimes  N^*\otimes N \otimes \left(N^*\otimes N \right)^*\\
\textrm{Spain} & np & N^*\otimes N\\
\end{array}\]
Given this lexicon, "tall person from Spain" has two derivations, corresponding to the bracketings
"(tall person) from Spain" ($x/\mathit{tall}, y/\mathit{person}, w/\mathit{from}, z/\mathit{Spain}$):
\[\scalebox{.9}{\prftree[straight][r]{$\backslash _{E_3}$}{\prftree[straight][r]{$/ _{E_2}$}{\prftree[straight][r]{$_{ax}$}{}{x:n/n \vdash x:n/n}}{\prftree[straight][r]{$_{ax}$}{}{y: n \vdash y:n}}{ (x:n/n, y:n) \vdash (x\triangleleft y) : n}}{\prftree[straight][r]{$/ _{E_1}$}{\prftree[straight][r]{$_{ax}$}{}{w:(n\backslash n)/np \vdash w:(n\backslash n)/np}}{\prftree[straight][r]{$_{ax}$}{}{z:np \vdash z:np}}{(w:(n\backslash n)/np, z:n) \vdash (w \triangleleft z):n\backslash n}}{[(x:n/n, y:n), (w:(n\backslash n)/np, z:n)] \vdash ((x\triangleleft y)\triangleright(w \triangleleft z)):n}
}\]
versus "tall (person from Spain)":
\[\scalebox{.9}{\prftree[straight][r]{$/ _{E_3}$}{\prftree[straight][r]{$_{ax}$}{}{x:n/n \vdash x:n/n}}{\prftree[straight][r]{$\backslash _{E_2}$}{\prftree[straight][r]{$_{ax}$}{}{y: n \vdash y:n}}{\prftree[straight][r]{$/ _{E_1}$}{\prftree[straight][r]{$_{ax}$}{}{w:(n\backslash n)/np \vdash w:(n\backslash n)/np}}{\prftree[straight][r]{$_{ax}$}{}{z:np \vdash z:np}}{(w:(n\backslash n)/np, z:n) \vdash (w \triangleleft z):n\backslash n}}{[y:n, (w:(n\backslash n)/np, z:n)] \vdash (y \triangleright (w \triangleleft z)):n}}{\left( x:n/n, [y:n, (w:(n\backslash n)/np, z:n)] \right) \vdash (x \triangleleft (y \triangleright (w \triangleleft z))):n}
}\] In the first reading, the adjective "tall" is evaluated with respect to all people, before it is specified that this person happens to be from Spain, whereas in the second reading the adjective "tall" is evaluated only in the restricted universe of people from Spain.

Taking "from Spain" as a unit for simplicity, let us start with the following primitive interpretations:

\renewcommand{\Spc}{}

\begin{itemize}
\item{$\llbracket tall^{n/n} \rrbracket_I=\sum_{ii',jj'} \prescript{}{}{\textbf{T}_{ii'}^{\; \;j'j}} \ket{^i_{\;j'}} \prescript{}{\Spc{N} \otimes \Spc{N}^*}{\bra{^{i'}_{\; j}}}$,}
\item $\llbracket person^{n} \rrbracket_I= \sum_{kk'} \prescript{}{}{\textbf{P}_{kk'}} \ket{^k} \prescript{}{\Spc{N}}{\bra{^{k'}}}$,
\item{$\llbracket from\_Spain^{n\backslash n} \rrbracket_I= \sum_{ll',mm'} \prescript{}{}{\textbf{F}_{\; \;mm'}^{l'l}} \ket{^{\;m}_{l'}} \prescript{}{\Spc{N}^* \otimes \Spc{N}}{\bra{^{\;m'}_{l}}}$.}
\end{itemize}

\begin{figure}
    \centering
    \includegraphics[scale=0.2]{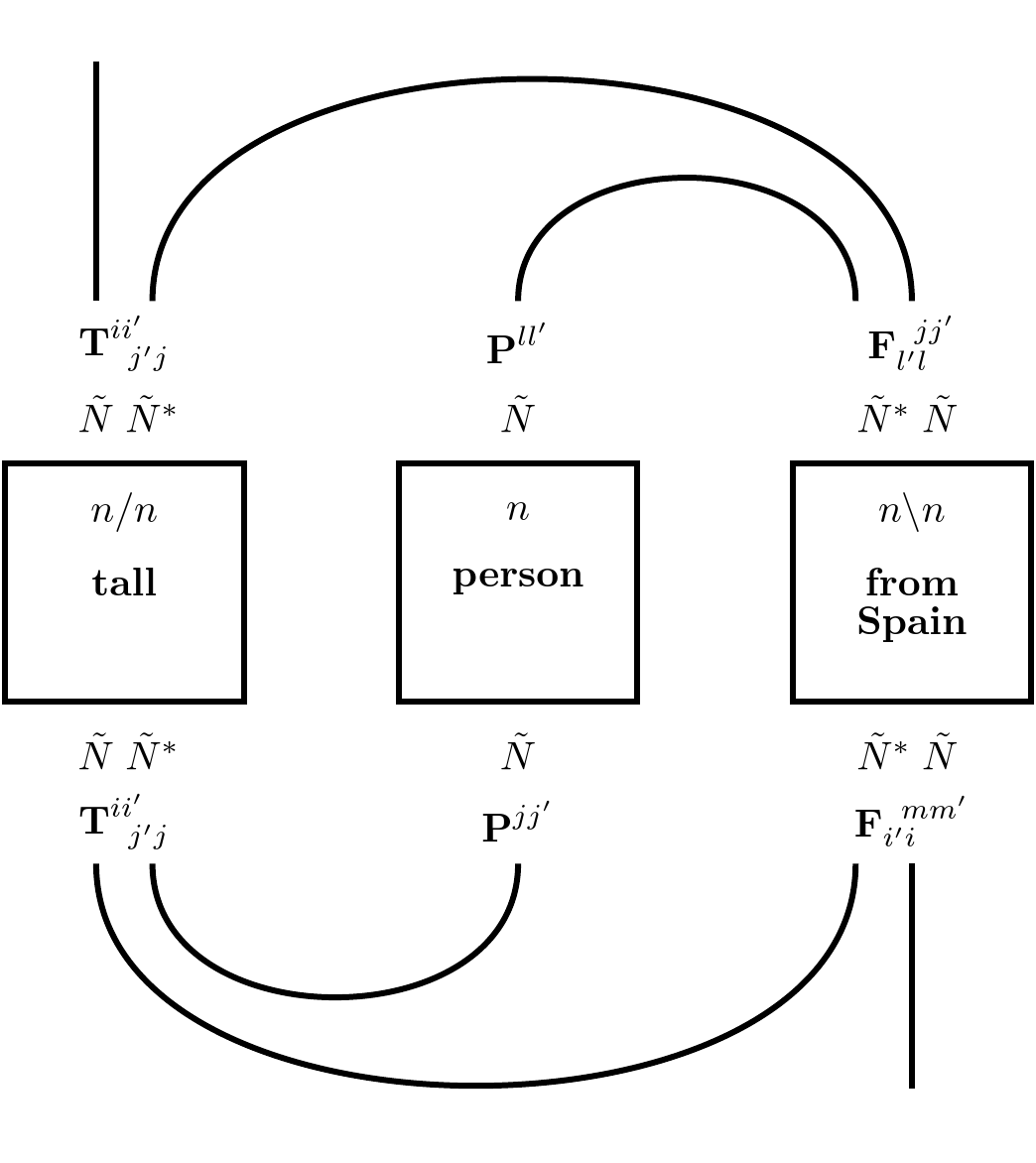}
    \caption{Representation of contractions corresponding to the first reading (lower links) and to the second reading (upper links), without subsystems. The final value is a coefficient in the $\tilde{N}$ space as in eq.\ref{first_reading_1} and in eq.\ref{second_reading_1}, respectively.}
    \label{fig:combo.1}
\end{figure}

By interpreting each step of the derivation in the way described in the previous section will give two different outcomes.The first one is

\begin{align}
&\llbracket tall\_person\_from\_Spain^{n} \rrbracket^1_I = \nonumber\\
=&\Tr_{\Spc{N}} \left( \Tr_{\Spc{N}} \left( \sum_{ii',jj'} \prescript{}{}{\textbf{T}^{ii'}_{\; \;j'j}} \ket{_i^{\;j'}} \prescript{}{\Spc{N} \otimes \Spc{N}^*}{\bra{_{i'}^{\; j}}} \cdot \sum_{kk'}  \prescript{}{}{\textbf{P}^{kk'}} \ket{_k} \prescript{}{\Spc{N}}{\bra{_{k'}}}   \right) \right. \nonumber \\
& \left. \cdot \sum_{ll',mm'} \prescript{}{}{\textbf{F}^{\; \;mm'}_{l'l}} \ket{_{\;m}^{l'}} \prescript{}{\Spc{N}^* \otimes \Spc{N}}{\bra{_{\;m'}^{l}}} \right) \nonumber \\
=& \sum_{ii',jj',mm'} \prescript{}{}{\textbf{T}^{ii'}_{\; \;j'j}}  \prescript{}{}{\textbf{P}^{jj'}} \prescript{}{}{\textbf{F}^{\; \;mm'}_{i'i}} \ket{_{m}} \prescript{}{\Spc{N}}{\bra{_{m'}}}, \label{first_reading_1}
\end{align} while the second one is

\begin{align}
&\llbracket tall\_person\_from\_Spain^{n} \rrbracket^2_I= \nonumber\\
=&\Tr_{\Spc{N}} \left( \sum_{ii',jj'} \prescript{}{}{\textbf{T}^{ii'}_{\; \;j'j}} \ket{_i^{\;j'}} \prescript{}{\Spc{N} \otimes \Spc{N}^*}{\bra{_{i'}^{\; j}}} \cdot \Tr_{\Spc{N}} \left(  \sum_{kk'}  \prescript{}{}{\textbf{P}_{kk'}} \ket{_k} \prescript{}{\Spc{N}}{\bra{_{k'}}}  \right. \right. \nonumber  \\
& \left. \left. \cdot \sum_{ll',mm'} \prescript{}{}{\textbf{F}^{\; \;mm'}_{l'l}} \ket{_{\;m}^{l'}} \prescript{}{\Spc{N}^* \otimes \Spc{N}}{\bra{_{\;m'}^{l}}} \right) \right) \nonumber \\
=& \sum_{ii',jj',ll'} \prescript{}{}{\textbf{T}^{ii'}_{\; \;j'j}} \prescript{}{}{\textbf{P}^{ll'}} \prescript{}{}{\textbf{F}^{\; \;jj'}_{l'l}} \ket{_{i}} \prescript{}{\Spc{N}}{\bra{_{i'}}}. \label{second_reading_1}
\end{align} The respective graphical representations of these contractions can be found in fig.\ref{fig:combo.1}.

\begin{figure}
    \centering
    \includegraphics[scale=0.2]{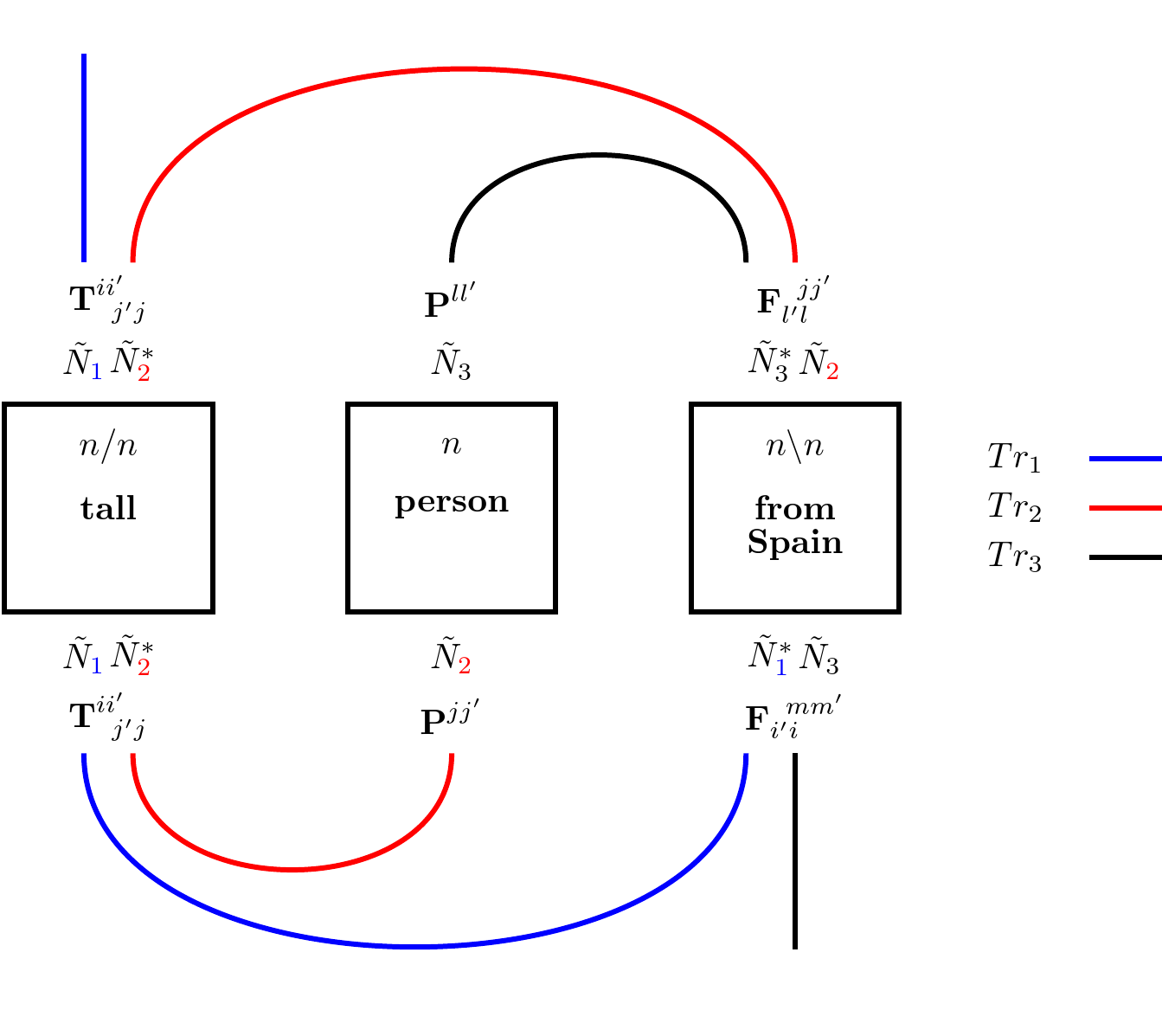}
    \caption{Representation of contractions corresponding to the first reading (lower links) and to the second reading (upper links), with subsystems. The final value is a coefficient in the $\tilde{N}$ space as in eq.\ref{first_reading} and in eq.\ref{second_reading}, respectively.}
    \label{fig:combo.2}
\end{figure}

\noindent Though the coefficients might be  different for each derivation, it is not clear how both interpretations are carried separately if they are part of a larger fragment, since their description takes place on the same space. Also, this recipe gives a fixed ordering and range for each trace. To be able to describe each final meaning separately, we use here the concept of  \textit{subsystem}. Because different subsystems act formally as different syntactic types and in each derivation the words that interact are different, it follows that each word should be assigned to a different subsystem:
\begin{itemize}[noitemsep]
\item{$\llbracket tall^{n/n} \rrbracket_{I_1} = \llbracket tall^{n/n} \rrbracket_{I_2} = \sum_{ii',jj'} \prescript{}{}{\textbf{T}^{ii'}_{\; \;j'j}} \ket{_i^{\;j'}} \prescript{}{\Spc{N}^1 \otimes \Spc{N}^2{^*}}{\bra{_{i'}^{\; j}}},$}
\item $\llbracket person^{n} \rrbracket_{I_1}= \sum_{kk'} \prescript{}{}{\textbf{P}^{kk'}} \ket{_k} \prescript{}{\Spc{N}^2}{\bra{_{k'}}}$, \\
$\llbracket person^{n} \rrbracket_{I_2}= \sum_{kk'} \prescript{}{}{\textbf{P}^{kk'}} \ket{_k} \prescript{}{\Spc{N}^3}{\bra{_{k'}}},$
\item{$\llbracket from\_Spain^{n\backslash n} \rrbracket_{I_1}= \sum_{ll',mm'} \prescript{}{}{\textbf{F}^{\; \;mm'}_{l'l}} \ket{_{\;m}^{l'}} \prescript{}{\Spc{N}^1{^*} \otimes \Spc{N}^3}{\bra{_{\;m'}^{l}}}$, \\
$\llbracket from\_Spain^{n\backslash n} \rrbracket_{I_2}= \sum_{ll',mm'} \prescript{}{}{\textbf{F}^{\; \;mm'}_{l'l}} \ket{_{\;m}^{l'}} \prescript{}{\Spc{N}^3{^*} \otimes \Spc{N}^2}{\bra{_{\;m'}^{l}}}$.}
\end{itemize} Notice that the value of the coefficients given by the interpretation functions $I_1$ and $I_2$ that describe the words does not change from the ones given in $I$, only possibly the subsystem assignment does.
Rewriting the derivation of the interpretations in terms of subsystems, the ordering of the traces does not matter anymore since the contraction is restricted to its own subsystem.
For the first reading we obtain

\begin{align}
&\llbracket tall\_person\_from\_Spain^{n} \rrbracket^{1}_{I_1} = \nonumber \\
=& \Tr_{\Spc{N}^1} \left( \Tr_{\Spc{N}^2} \left( \sum_{ii',jj'} \prescript{}{}{\textbf{T}^{ii'}_{\; \;j'j}} \ket{_i^{\;j'}} \prescript{}{\Spc{N}^1 \otimes \Spc{N}^2{^*}}{\bra{_{i'}^{\; j}}} \cdot \sum_{kk'} \prescript{}{}{\textbf{P}^{kk'}} \ket{_k} \prescript{}{\Spc{N}^2}{\bra{_{k'}}}  \right. \right. \nonumber \\
& \left. \left.  \cdot\sum_{ll',mm'}  \prescript{}{}{\textbf{F}^{\; \;mm'}_{l'l}} \ket{_{\;m}^{l'}} \prescript{}{\Spc{N}^1{^*} \otimes \Spc{N}^3}{\bra{_{\;m'}^{l}}} \right) \right) \nonumber \\
=& \sum_{ii',jj',mm'} \prescript{}{}{\textbf{T}^{ii'}_{\; \;j'j}} \prescript{}{}{\textbf{P}^{jj'}} \prescript{}{}{\textbf{F}^{\; \;mm'}_{i'i}} \ket{_{m}} \prescript{}{\Spc{N}^3}{\bra{_{m'}}} \label{first_reading}
\end{align} and for the second

\begin{align}
&\llbracket tall\_person\_from\_Spain^{n} \rrbracket^{2}_{I_2}= \nonumber \\
=&\Tr_{\Spc{N}^2} \left( \sum_{ii',jj'} \prescript{}{}{\textbf{T}^{ii'}_{\; \;j'j}} \ket{_i^{\;j'}} \prescript{}{\Spc{N}^1 \otimes \Spc{N}^2{^*}}{\bra{_{i'}^{\; j}}} \cdot \Tr_{\Spc{N}^3} \left(  \sum_{kk'}  \prescript{}{}{\textbf{P}^{kk'}} \ket{_k} \prescript{}{\Spc{N}^3}{\bra{_{k'}}} \right. \right. \nonumber \\
& \left. \left. \cdot \sum_{mm',ll'} \prescript{}{}{\textbf{F}^{\; \;mm'}_{l'l}} \ket{_{\;m}^{l'}} \prescript{}{\Spc{N}^3{^*} \otimes \Spc{N}^2}{\bra{_{\;m'}^{l}}} \right) \right) \nonumber \\
=& \Tr_{\Spc{N}^3} \left( \Tr_{\Spc{N}^2} \left( \sum_{ii',jj'} \prescript{}{}{\textbf{T}^{ii'}_{\; \;j'j}} \ket{_i^{\;j'}} \prescript{}{\Spc{N}^1 \otimes \Spc{N}^2{^*}}{\bra{_{i'}^{\; j}}} \cdot  \sum_{kk'} \prescript{}{}{\textbf{P}^{kk'}} \ket{_k} \prescript{}{\Spc{N}^3}{\bra{_{k'}}}   \right. \right. \nonumber \\
& \left. \left. \cdot \sum_{ll',mm'}  \prescript{}{}{\textbf{F}^{\; \;mm'}_{l'l}} \ket{_{\;m}^{l'}} \prescript{}{\Spc{N}^3{^*} \otimes \Spc{N}^2}{\bra{_{\;m'}^{l}}} \right) \right) \nonumber \\
=& \sum_{ii',jj',ll'} \prescript{}{}{\textbf{T}^{ii'}_{\; \;j'j}} \prescript{}{}{\textbf{P}^{ll'}} \prescript{}{}{\textbf{F}^{\; \;jj'}_{l'l}} \ket{_{i}} \prescript{}{\Spc{N}^1}{\bra{_{i'}}}. \label{second_reading}
\end{align} The interpretation of each derivation belongs now to different subsystems, which keeps the information about the original word to which the free "noun" space is attached. We can see this by comparing the upper and lower links in fig. \ref{fig:combo.2}.

\begin{figure}
    \centering
    \includegraphics[scale=0.09]{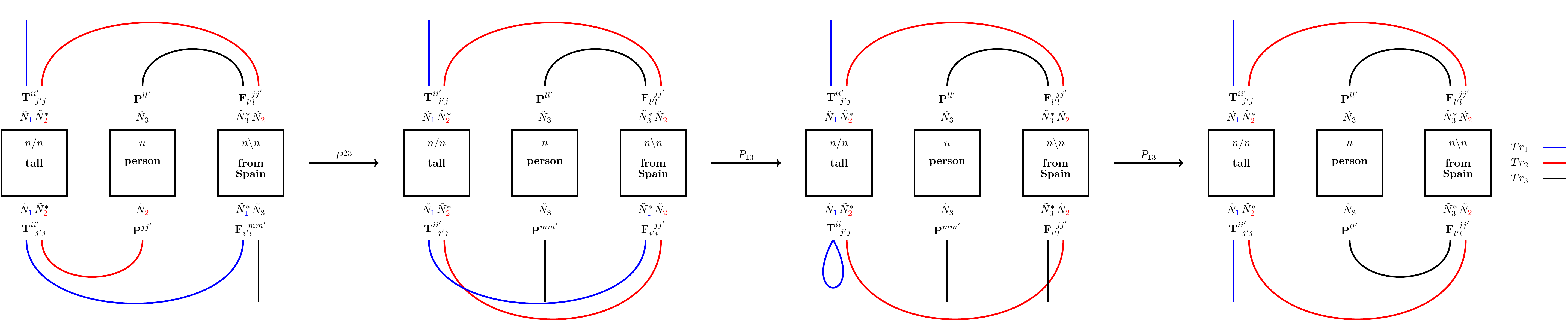}
    \caption{}
    \label{fig:prog.1}
\end{figure}

\noindent However, it is not very convenient to attribute each word to a different subsystem depending on the interpretation it will be part of, since that is information that comes from the derivation itself and not from the representations of words. To tackle this problem, one uses permutation operations over the subsystems. Since these have precedence over the trace, when the traces are taken the contractions change accordingly. This changes the subsystem assignment at specific points so it is possible to go from one interpretation to the other, without giving different interpretations to each word initially. Thus, there is a way to go directly from the first interpretation to the second:

\begin{align}
&\llbracket tall\_person\_from\_Spain^{n}\rrbracket^2_{I_1} = \nonumber \\
=& \Tr_{\Spc{N}^1} \left( P_{13} \Tr_{\Spc{N}^2} \left( \sum_{ii',jj'} \prescript{}{}{\textbf{T}^{ii'}_{\; \;j'j}} \ket{_i^{\;j'}} \prescript{}{\Spc{N}^1 \otimes \Spc{N}^2{^*}}{\bra{_{i'}^{\; j}}} \cdot P_{13} P^{23}   \sum_{kk'} \prescript{}{}{\textbf{P}^{kk'}} \ket{_k} \prescript{}{\Spc{N}^2}{\bra{_{k'}}}  \right.  \right. \nonumber \\
& \left. \left. \cdot  \sum_{ll',mm'} \prescript{}{}{\textbf{F}^{\; \;mm'}_{l'l}} \ket{_{\;m}^{l'}} \prescript{}{\Spc{N}^1{^*} \otimes \Spc{N}^3}{\bra{_{\;m'}^{l}}} P^{23}  P_{13} \right) P_{13} \right) \nonumber\\
=& \Tr_{\Spc{N}^3} \left( \Tr_{\Spc{N}^2} \left(\sum_{ii',jj'} \prescript{}{}{\textbf{T}^{ii'}_{\; \;j'j}} \ket{_i^{\;j'}} \prescript{}{\Spc{N}^1 \otimes \Spc{N}^2{^*}}{\bra{_{i'}^{\; j}}} \cdot \sum_{kk'}  \prescript{}{}{\textbf{P}^{kk'}} \ket{_k} \prescript{}{\Spc{N}^3}{\bra{_{k'}}}   \right. \right. \nonumber \\ 
& \left. \left. \cdot \sum_{ll',mm'} \prescript{}{}{\textbf{F}^{\; \;mm'}_{l'l}} \ket{_{\;m}^{l'}} \prescript{}{\Spc{N}^3{^*} \otimes \Spc{N}^2}{\bra{_{\;m'}^{l}}} \right) \right).
\end{align} The reasoning behind is as follows: the permutation $P^{23}$ swaps the space assignment between "person" and the free space in "from\_Spain", according to eq.\ref{permup}; after that a permutation $P_{13}$ is used as in eq. \ref{permone} to change the argument space of "from\_Spain" from $\Spc{N}^{1{^*}}$ to $\Spc{N}^{3{^*}}$, and then the same permutation is applied again to change the space of tracing, following eq.\ref{permtrace}. In this way, all the coefficients will have the correct contractions and in a different space from the first reading. In fig. \ref{fig:prog.1} we can see the action of the permutations by visualizing how both the spaces and the traces change as we go from the lower to the upper links.

\noindent Although the metric is not used explicitly in the application of the permutation operators, it is necessary to generate the correct tensors where the permutation operator is applied in the first place, by going from the vector representation that comes directly from the data to one that allows contraction. As an example, the adjective "tall" would have a vector representation from the data as an element of $\tilde{V} \otimes \tilde{V}$, of the form $\textbf{T}^{ii', kk'}$. We need the metric $d_{kj'}d_{k'j}$ to change its form to $T^{ii'}_{j'j}$. By defining the interpretation space of adjectives as $\Tilde{N} \otimes \Tilde{N}^*$, we assume this passage has already been made when we assign an interpretation to a word in this space. As an alternative to this derivation, we mention that it is possible to apply a $P^{23}$ permutation followed by a $P^{13}$ permutation that results in the correct contraction of the indices, but fails to deliver the results of the two derivations in different subspaces; it is however noteworthy that, in order to start with a unique assignment for each word, this alternative derivation can, in any case, only be achieved by distinguishing between subsystems, as well as the covariant and contravariant indices.

\nocite{moot2012logic}

\nocite{moortgat1997categorial}

\section{Conclusion and Future Work}

In this paper we provided a density matrix model for a simple fragment of the Lambek Calculus, differently from what is done in \cite{blacoe2014semantic} who uses density matrices to interpret dependency parse trees. The syntax-semantics interface takes the form of a compositional map assigning semantic values to the $\lambda_{/,\bs}$ terms coding syntactic derivations. We proposed the use of a metric as a way to reconcile the various vector representations of the same word that come from different treatments, assuming that there is a quantity that is being preserved, like human judgements. If we know the metric, we can confidently assign only one embedding to each word as its semantic value. A metric can relate these various representations so that we can assign only one vector as its semantic value. The density matrix model enables the integration of lexical and derivational forms of ambiguity. Additionaly, it allows for the transfer of methods and techniques from quantum mechanics and general relativity to computational semantics. An example of such transfer is the permutation operator. In quantum mechanics, this operator permits a description of indistinguishable particles. In the linguistic application, it allows one to go from an interpretation that comes from one derivation to another, without the need to to go through the latter, but keeping this second meaning in a different subsystem. Another example is the introduction of covariant and contravariant components, associated with a metric, that allow the permutation operations to be properly applied. In future work, we want to explore the preservation of human judgements found in the literature via a metric that represents the variability of vector representations of words, either static or dynamic. We also want to extend our simple fragment with modalities for structural control (cf \cite{moortgat1997categorial}), in order to deal with cases of derivational ambiguity that are licensed by these control modalities. Finally, we want to consider derivational ambiguity in the light of an \emph{incremental} left-to-right interpretation process, so as to account for the evolution of interpretations over time. In enriching the treatment with a metric, we want to explore the consequences of having this new parameter in treating context dependent embeddings.

\section{Acknowledgements}

A.D.C. thanks discussions with Sanjaye Ramgoolam and Martha Lewis that contributed to the duality concepts included in the journal version of this paper. This work is supported by the Complex Systems Fund, with special thanks to Peter Koeze.

\bibliographystyle{plain}
\bibliography{bib}

\begin{thebibliography}{10}

\bibitem{bankova2019graded}
Dea Bankova, Bob Coecke, Martha Lewis, and Dan Marsden.
\newblock Graded hyponymy for compositional distributional semantics.
\newblock {\em Journal of Language Modelling}, 6(2):225--260, 2019.

\bibitem{blacoe2014semantic}
William Blacoe.
\newblock Semantic composition inspired by quantum measurement.
\newblock In {\em International Symposium on Quantum Interaction}, pages
  41--53. Springer, 2014.

\bibitem{bruni2014multimodal}
Elia Bruni, Nam-Khanh Tran, and Marco Baroni.
\newblock Multimodal distributional semantics.
\newblock {\em Journal of Artificial Intelligence Research}, 49:1--47, 2014.

\bibitem{clark2015vector}
Stephen Clark.
\newblock Vector space models of lexical meaning.
\newblock {\em Handbook of Contemporary Semantics}, 10:9781118882139, 2015.

\bibitem{coecke2010mathematical}
Bob Coecke, Mehrnoosh Sadrzadeh, and Stephen Clark.
\newblock Mathematical foundations for a compositional distributional model of
  meaning.
\newblock {\em Lambek Festschrift, Linguistic Analysis 36(1--4)}, pages
  345--384, 2010.

\bibitem{correia2020putting}
AD~Correia, HTC Stoof, and M~Moortgat.
\newblock Putting a spin on language: A quantum interpretation of unary
  connectives for linguistic applications.
\newblock {\em arXiv preprint arXiv:2004.04128}, 2020.

\bibitem{devlin-etal-2019-bert}
Jacob Devlin, Ming-Wei Chang, Kenton Lee, and Kristina Toutanova.
\newblock {BERT}: Pre-training of deep bidirectional transformers for language
  understanding.
\newblock In {\em Proceedings of the 2019 Conference of the North {A}merican
  Chapter of the Association for Computational Linguistics: Human Language
  Technologies, Volume 1 (Long and Short Papers)}, pages 4171--4186,
  Minneapolis, Minnesota, June 2019. Association for Computational Linguistics.

\bibitem{dullemond1991introduction}
Kees Dullemond and Kasper Peeters.
\newblock Introduction to tensor calculus.
\newblock {\em Kees Dullemond and Kasper Peeters}, 1991.

\bibitem{georgi1982lie}
Howard Georgi.
\newblock Lie algebras in particle physics. from isospin to unified theories.
\newblock {\em Front. Phys.}, 54:1--255, 1982.

\bibitem{grefenstette2011experimental}
Edward Grefenstette and Mehrnoosh Sadrzadeh.
\newblock Experimental support for a categorical compositional distributional
  model of meaning.
\newblock In {\em Proceedings of the Conference on Empirical Methods in Natural
  Language Processing}, pages 1394--1404. Association for Computational
  Linguistics, 2011.

\bibitem{harris1954distributional}
Zellig~S Harris.
\newblock Distributional structure.
\newblock {\em Word}, 10(2-3):146--162, 1954.

\bibitem{LoLacompositionality}
Theo~M.V. Janssen and Barbara~H. Partee.
\newblock Chapter 7 - {C}ompositionality.
\newblock In Johan van Benthem and Alice ter Meulen, editors, {\em Handbook of
  Logic and Language}, pages 417 -- 473. North-Holland, Amsterdam, 1997.

\bibitem{lambek1958mathematics}
Joachim Lambek.
\newblock The mathematics of sentence structure.
\newblock {\em The American Mathematical Monthly}, 65(3):154--170, 1958.

\bibitem{lam61}
Joachim Lambek.
\newblock On the calculus of syntactic types.
\newblock In Roman Jakobson, editor, {\em Structure of Language and its
  Mathematical Aspects}, volume XII of {\em Proceedings of Symposia in Applied
  Mathematics}, pages 166--178. American Mathematical Society, 1961.

\bibitem{mikolov2013distributed}
Tomas Mikolov, Ilya Sutskever, Kai Chen, Greg~S Corrado, and Jeff Dean.
\newblock Distributed representations of words and phrases and their
  compositionality.
\newblock In {\em Advances in neural information processing systems}, pages
  3111--3119, 2013.

\bibitem{mitchell2010composition}
Jeff Mitchell and Mirella Lapata.
\newblock Composition in distributional models of semantics.
\newblock {\em Cognitive science}, 34(8):1388--1429, 2010.

\bibitem{moortgat1997categorial}
Michael Moortgat.
\newblock Chapter 2 - {C}ategorial type logics.
\newblock In Johan van Benthem and Alice ter Meulen, editors, {\em Handbook of
  Logic and Language}, pages 93--177. Elsevier, Amsterdam, 1997.

\bibitem{moot2012logic}
Richard Moot and Christian Retor{\'e}.
\newblock {\em The logic of categorial grammars: a deductive account of natural
  language syntax and semantics}, volume 6850.
\newblock Springer, 2012.

\bibitem{nielsen2002quantum}
Michael~A Nielsen and Isaac Chuang.
\newblock Quantum computation and quantum information, 2002.

\bibitem{pennington2014glove}
Jeffrey Pennington, Richard Socher, and Christopher~D Manning.
\newblock Glove: Global vectors for word representation.
\newblock In {\em Proceedings of the 2014 conference on empirical methods in
  natural language processing (EMNLP)}, pages 1532--1543, 2014.

\bibitem{peters-etal-2018-deep}
Matthew Peters, Mark Neumann, Mohit Iyyer, Matt Gardner, Christopher Clark,
  Kenton Lee, and Luke Zettlemoyer.
\newblock Deep contextualized word representations.
\newblock In {\em Proceedings of the 2018 Conference of the North {A}merican
  Chapter of the Association for Computational Linguistics: Human Language
  Technologies, Volume 1 (Long Papers)}, pages 2227--2237, New Orleans,
  Louisiana, June 2018. Association for Computational Linguistics.

\bibitem{piedeleu2014ambiguity}
Robin Piedeleu.
\newblock {\em Ambiguity in categorical models of meaning}.
\newblock PhD thesis, University of Oxford Master’s thesis, 2014.

\bibitem{DBLP:journals/corr/PiedeleuKCS15}
Robin Piedeleu, Dimitri Kartsaklis, Bob Coecke, and Mehrnoosh Sadrzadeh.
\newblock Open system categorical quantum semantics in natural language
  processing.
\newblock {\em CoRR}, abs/1502.00831, 2015.

\bibitem{sadrzadeh2018sentence}
Mehrnoosh Sadrzadeh, Dimitri Kartsaklis, and Esma Balk{\i}r.
\newblock Sentence entailment in compositional distributional semantics.
\newblock {\em Annals of Mathematics and Artificial Intelligence},
  82(4):189--218, 2018.

\bibitem{sadun2007applied}
Lorenzo~Adlai Sadun.
\newblock {\em Applied linear algebra: The decoupling principle}.
\newblock American Mathematical Soc., 2007.

\bibitem{selinger2007dagger}
Peter Selinger.
\newblock Dagger compact closed categories and completely positive maps.
\newblock {\em Electronic Notes in Theoretical computer science}, 170:139--163,
  2007.

\bibitem{vbenthem1983}
Johan van Benthem.
\newblock The semantics of variety in categorial grammar.
\newblock Technical Report 83-29, Simon Fraser University, Burnaby (B.C.),
  1983.
\newblock Revised version in W. Buszkowski, W. Marciszewski and J. van Benthem
  (eds) Categorial grammar, Benjamin, Amsterdam.

\bibitem{wald1984general}
Robert~M Wald.
\newblock General relativity.
\newblock {\em University of Chicago Press}, 1984.

\bibitem{wansing1992}
Heinrich Wansing.
\newblock Formulas-as-types for a hierarchy of sublogics of intuitionistic
  propositional logic.
\newblock In David Pearce and Heinrich Wansing, editors, {\em Nonclassical
  Logics and Information Processing}, pages 125--145, Berlin, Heidelberg, 1992.
  Springer Berlin Heidelberg.

\bibitem{wijnholds2019evaluating}
Gijs Wijnholds and Mehrnoosh Sadrzadeh.
\newblock Evaluating composition models for verb phrase elliptical sentence
  embeddings.
\newblock In {\em Proceedings of the 2019 Conference of the North American
  Chapter of the Association for Computational Linguistics: Human Language
  Technologies, Volume 1 (Long and Short Papers)}, pages 261--271, 2019.

\end{thebibliography}

\newpage
\appendix

\end{document}